\documentclass[11pt,a4paper]{article}
\usepackage[hyperref]{emnlp2020}
\usepackage{times}
\usepackage{latexsym}

\usepackage{microtype}
\usepackage{booktabs} 
\usepackage{breakurl}

\usepackage{amsmath}
\usepackage{graphicx}
\usepackage{caption}
\usepackage{subcaption}
\usepackage{algorithm}
\usepackage{algorithmic}

\aclfinalcopy 

\title{Rank and run-time aware compression of NLP Applications}

\author{Urmish Thakker \\
  SambaNova Systems \\ 
  \texttt{uthakker@cs.wisc.edu} \\ \\
  \textbf{Ganesh Dasika} \\
  AMD Research \\
  \And
  \textbf{Jesse Beu} \\ 
  Arm ML Research \\  \\ \\
  \textbf{Matthew Mattina} \\
  Arm ML Research \\ \And
  Dibakar Gope \\
  Arm ML Research \\ }

\date{}

\begin{document}
\maketitle
\maketitle

\begin{abstract}
Sequence model based NLP applications can be large. Yet, many applications that benefit from them run on small devices with very limited compute and storage capabilities, while still having run-time constraints. As a result, there is a need for a compression technique that can achieve significant compression without negatively impacting inference run-time and task accuracy. This paper proposes a new compression technique called Hybrid Matrix Factorization that achieves this dual objective. HMF improves low-rank matrix factorization (LMF) techniques by doubling the rank of the matrix using an intelligent hybrid-structure leading to better accuracy than LMF. Further, by preserving dense matrices, it leads to faster inference run-time than pruning or structure matrix based compression technique. We evaluate the impact of this technique on 5 NLP benchmarks across multiple tasks (Translation, Intent Detection, Language Modeling) and show that for similar accuracy values and compression factors, HMF can achieve more than $2.32\times$ faster inference run-time than pruning and 16.77\% better accuracy than LMF.
\end{abstract}

\section{Introduction}
\label{intro}
Sequence based (LSTMs/GRUs) NLP Applications are being increasingly run on mobile phones and smart watches. They are typically enabled by querying a cloud-based system to do most of the computation. The energy, latency, and privacy implications associated with running a query on the cloud is changing where users run a neural network application. We should, therefore, expect an increase in the number of NLP applications running on embedded devices. Due to the energy and power constraints of edge devices, embedded SoCs frequently use lower-bandwidth memory technologies and smaller caches compared to desktop and server processors. Thus, there is a need for good compression techniques to enable large NLP models to fit into an smaller edge device or ensure that they run efficiently on devices with smaller caches \cite{urmtha01RNN}. Additionally, compressing models should not negatively impact the inference run-time as these tasks may have real-time deadlines to provide a good user experience. 

In order to choose a compression scheme for a particular network, one needs to consider 3 different axes -- the compression factor, the inference run-time speedup over the baseline, and the accuracy. Ideally, a good compression algorithm should not sacrifice improvement along one axis for improvement along another. For example, network pruning~\cite{han2015deep_compression} has shown to be an effective compression technique, but pruning creates a sparse matrix representation that is inefficient to execute on most modern CPUs. Our analysis shows that pruned networks can achieve a faster run-time than the baseline only for significantly high compression factors. Low-rank matrix factorization (LMF) is another popular compression technique that can achieve speedup proportional to the compression factor. However, LMF has had mixed results in maintaining model accuracy~\cite{lmf-good1,lmf-bad,DBLP:journals/corr/LuSS16}. This is because LMF reduces the rank of a matrix significantly, reducing its expressibility \cite{rnn-rank}. Lastly, structured matrices \cite{structuredmatrix} can also be used to compress neural networks. While these techniques show a significant reduction in computation, this reduction only translates to a realized run-time improvement for large matrices ~\cite{NIPS2018_8119} or while using specialized hardware~\cite{clstm-fpga,circular2}. For benchmarks evaluated in this paper, HMF gets $30\times$ speed-up improvement over structured matrix based technique \cite{circular2}.

Given LMF's good run-time characteristics, it can potentially act as an alternative to pruning. However, LMF leads to an accuracy loss. \textbf{To overcome the problem of finding an alternative to pruning, which preserves the run-time benefit of dense structures of LMF and the accuracy benefits of pruned networks, we introduce a new compression technique called Hybrid Matrix Factorization (HMF). HMF can act as an effective compression technique for NLP edge use cases on embedded CPUs.} The results are very promising -- HMF achieves iso-accuracy for a large compression factor (2$\times$ to 4$\times$), improves the CPU run-time over pruning by a factor of $2.32\times$ and can achieve $16.77\%$ better model accuracy than LMF and 9\% better accuracy than smaller baselines. 


\section{Related Work}

\textbf{Pruning} \cite{han2015deep_compression,suyog,sanh2020movement} has been the most successful compression technique for all types of neural networks. Poor hardware characteristics of pruning has led to research in block based pruning technique  \cite{blockprune}. However, block based pruning technique also requires certain amount of block sparsity to achieve faster run-time than baseline. Having a strict compression factor requirement to get better run-time is a stringent constraint that HMF manages to avoid. 

\textbf{Structured matrices} have shown significant potential for compression of NN \cite{circular2,structuredmatrix,clstm,circnn,circular1,dkp}. Block circular compression is an extension of structured matrix based compression technique, converting every block in a matrix into structured matrix. We will show in this paper that HMF is a superior technique than block circular decomposition. 

\textbf{Tensor decomposition} (CP decomposition, Kronecker, Tucker decomposition etc) based methods have also shown significant reduction in parameters \cite{tjandra2017compressing,Thakker2019PushingTL,urmishKron}. Matrix Factorization \cite{DBLP:journals/corr/KuchaievG17,lmf-bad,lmf-good1,urmtha01-hmd} can be categorized under this topic. We will show in this paper that HMF can lead to better accuracy than LMF compressed RNNs. 

\textbf{Quantization} is another popular technique for compression \cite{Quant-hubara,Quant-bengio,Gope2019TernaryMV,sanh2019distilbert,bin-lstm,TernaryMobileNetstinyMLSummit2020,GopeMLSys2019}. Networks compressed using HMF can be further compressed using quantization.

\textbf{Dynamic techniques} are used to improve inference run-time of RNNs by skipping certain RNN state updates \cite{skiprnn,skimrnn,lstmjump,urmSkipRNN}. These techniques are based on the assumption that not all inputs to a RNN are needed for final classification task. Thus we can learn a small and fast predictor that can learn to skip certain inputs and its associated computation. HMF technique is orthogonal to this technique and networks compressed using HMF can be further optimized using this technique. 

\textbf{Design of efficient structures} for LSTM/GRU cells like SRU \cite{sru}, QRNN \cite{qrnn} and PRU \cite{pru} have also led to networks with faster inference run-time benefits or lesser number of parameters. These structures are different from structured matrices and are hand-crafted after better understanding the application domain. HLF can be further used to optimize the matrices in these architectures to make the resultant network more parameter and run-time efficient.

Finally, any technique used to reduce the parameter footprint of embedding matrices in NLP can further optimize RNN networks optimized using HMF  \cite{embed1,embed2}. In this paper, we show that HMF can compress networks with compressed word embedding layers.

\section{Hybrid Matrix Factorization}
\label{sec:hybrid}

\subsection{Why LMF can potentially lead to loss in accuracy}
 LMF ~\cite{DBLP:journals/corr/KuchaievG17} expresses a larger matrix $\boldsymbol{A} \in R^{m\times n}$ as a product of two smaller matrices $\boldsymbol{U}$ $\in R^{m\times r}$ and V $\in R^{r\times n}$, respectively. Parameter $r$ controls the compression factor.  Unlike pruning, matrix factorization is able to improve the run-time over the baseline for most compression factors. Unfortunately, compression via LMF can lead to loss in accuracy. We believe, this is because of two closely related reasons: 
 \begin{itemize}
     \item \textbf{Rank-Loss}: The rank of a matrix is a measure of the expressibility of a matrix. A lower rank matrix means less expressibility, limiting its learning capacity. This can potentially lead to some accuracy loss.  LMF compression leads to a lower ranked matrix. While before compression, the rank of matrix \textbf{A} is $min(m,n)$, after compression, it becomes $min(m,n,d)$.  Eg - If A $\in R^{256\times 256}$, compression using LMF by a factor of $2$ leads to $\boldsymbol{U}$ $\in R^{256\times 64}$ and $\boldsymbol{V}$ $\in R^{64\times 256}$. The resultant compressed matrix $\boldsymbol{A}$ (= $\boldsymbol{U}*\boldsymbol{V}$) is a 64 rank matrix. Thus, in order to compress the matrix by a factor of 2, LMF reduces the rank of a matrix by a factor of 4. 
     \item \textbf{Less expressive output features}: A closely related argument can be viewed when we extend the idea of low-rank matrix and its impact on the output features. Without loss of generality, an LSTM/GRU layer calculates a matrix-vector product during inference. If we assume the parameters of a LSTM/GRU layer are represented by a matrix $\boldsymbol{A}$ $\in R^{m\times n}$ and the input to the matrix is $x \in R^{n\times 1}$, then the output feature calculated is -
     \begin{gather*}
     y = f(\boldsymbol{A}*x) \\
     where,\; y \in R^{m\times 1}
     \end{gather*}. f is a non-linear function. Thus, each element of y is a dot product of a row of $\boldsymbol{A}$ and the vector $x$ followed by non-linearity. LMF expresses $\boldsymbol{A}$ in a lower dimensional space using the $\boldsymbol{U}$ and $\boldsymbol{V}$ matrix. If we rewrite the equation to calculate y, when \textbf{A} is expressed using LMF, we get -
     \begin{gather}
         y = f(\boldsymbol{\boldsymbol{U}}* \boldsymbol{\boldsymbol{V}}* x) \\
         y = f(\boldsymbol{U}* k)\; \; (assuming\; k = \boldsymbol{V}* x) \label{eq:lowdim} \\
          where,\; k \in R^{r\times 1}
     \end{gather}
     Generally, for compression $r < m,n$. Thus, x $\in R^{n\times 1}$ is projected to a lower dimensional embedding of size $R^{d\times 1}$ and expanded again to $R^{m\times 1}$ to create y. Thus, compressing \textbf{A} to a lower rank leads to output features calculated from a lower dimension embedding vector. 
 \end{itemize}

\subsection{Hybrid Matrix Factorization}
This paper introduces a new compression technique that uses dense matrix representation to ensure fast run-time properties and avoids making the strong assumptions made by LMF. This technique is based on three assumptions -
\begin{itemize}
    \item \textbf{A1:} Rank of a matrix is important to create a high-task accuracy LSTM/GRU network \cite{rnn-rank}
    \item \textbf{A2:} LMF makes a strong assumption that all the elements of the output feature vector of a LSTM/GRU Cell can be expressed from a low-dimensional embedding vector (Equation \ref{eq:lowdim}). HLF is based on the assumption that a more relaxed constraint of having a hybrid output feature vector, where some elements are calculated from a lower dimensional embedding space and other's from a higher dimensional embedding space can lead to better accuracy. 
    \item \textbf{A3:} Most LSTM/GRU networks are followed by a fully-connected softmax layer or another LSTM/GRU layer. Even if the order of the elements in the output of a particular RNN layer changes, the weights in the subsequent fully connected or LSTM/GRU layers can adjust to accommodate that. Thus, the order of the elements of the output vector of LSTM/GRU layer is not strictly important.
\end{itemize}
 
These three intuitions of a LSTM/GRU layer can be used to create a more hardware-friendly compression scheme. This paper introduces one such scheme -- Hybrid Matrix Factorization.
\begin{figure}[tb]
\vskip 0.1in
\centering
\includegraphics[width=0.7\columnwidth]{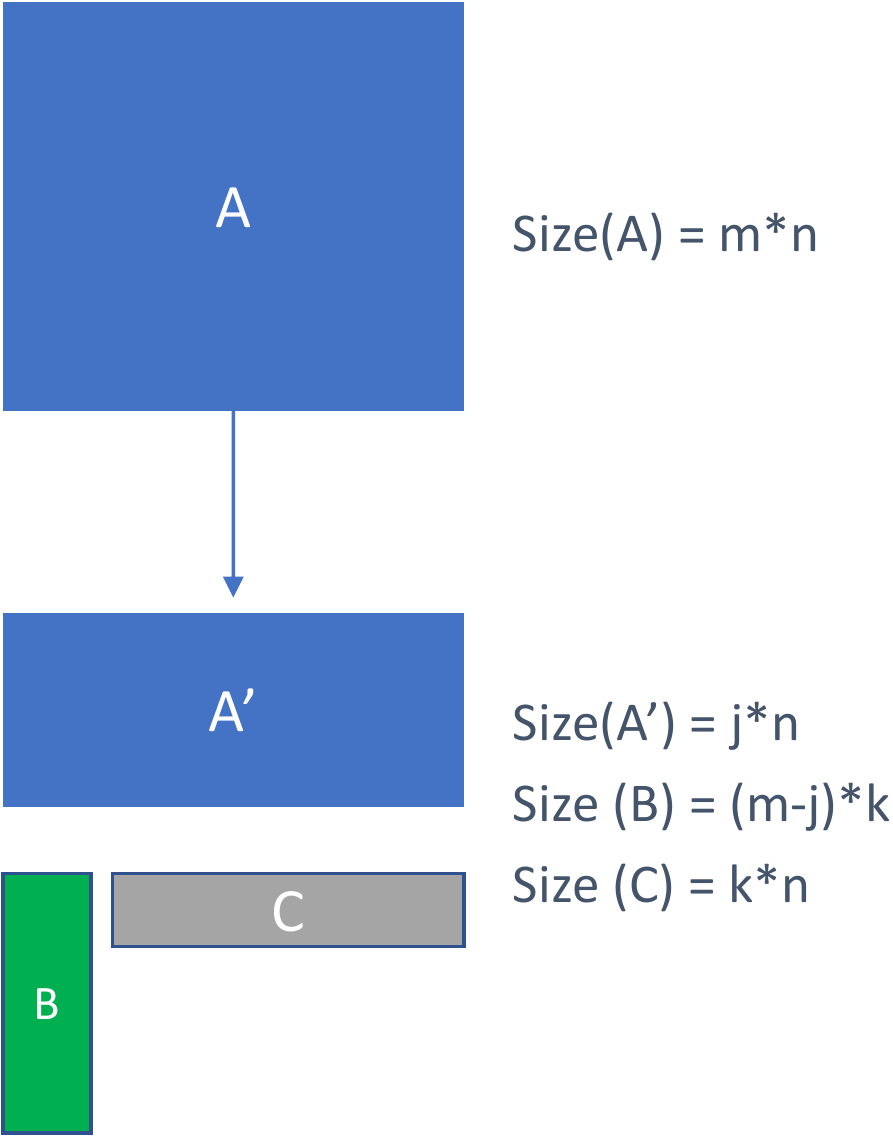}
\caption{Representation of a matrix using hybrid factorization}
\label{hmatrix-fig}
\vskip -0.1in
\end{figure}

\begin{algorithm}[t]
   {\bfseries Input 1:} Matrices $A'$ of dimension $j \times n$, $B$ of dimension $(m-j) \times k$, $C$ of dimension $k \times n)$\\
   {\bfseries Input 2:} Vector $I$ of dimension $n \times 1$ \\
   {\bfseries Output:} Matrix $O$ of dimension $m\times1$ \\
   \begin{algorithmic}[1]
   \STATE $O\textsubscript{1:j} \gets A'\times I$
   \STATE $Temp1 \gets C \times I$
   \STATE $O\textsubscript{j+1:m} \gets B \times Temp1$
   \STATE $O = concatenate\{O\textsubscript{1:j},O\textsubscript{j+1:m}\} $
\end{algorithmic}
\caption{Matrix vector product when a matrix uses the HLF technique}
\label{alg:hybrid-mvalg}
\end{algorithm}

Hybrid Matrix Factorization (HLF) splits the input and recurrent matrices in an LSTM/GRU layer into two parts -- a fully parameterized upper part and low-rank lower part. 

Figure \ref{hmatrix-fig} shows the strategy we use to decompose the matrix - an unconstrained upper half $\boldsymbol{A'}$ and a lower half that is composed of $k$ rank-1 blocks. If we decompose the weight matrix using this technique, the parameter reduction is given by:
\begin{equation}
\label{eq:HLF-d2}
\frac{m \times n}{(j \times n) + k\times(m-j + n)}
\end{equation}
Thus, the maximum rank of the matrix becomes $j+k$. Different values of j and k can be used to control the amount of compression and the rank of the matrix.

\begin{table}[t]
\centering
\begin{tabular}{lcc}
\toprule
\multicolumn{3}{c}{Matrix of Size (256,256)} \\
\toprule
Compression Factor  & LMF  & HLF        \\ 
\midrule
1.25                & 102  & 103 - 204  \\ 
1.67                & 76   & 78 - 153   \\ 
2.50                & 51   & 52 - 101   \\ 
5.00                & 25   & 26 - 50     \\ 
\bottomrule
\end{tabular}
\caption{The maximum possible rank of a $256\times 256$ sized matrix after it is compressed by 4 different factors using 3 different compression techniques. To compress a matrix by a given compression factor, HLF has 2 different parameters $j$ and $k$ to regulate the rank of the matrix. Hence, we see a range of rank values. Maximum rank is achieved when k=1. The value for j when k=1 can be calculated for different compression factors using equation \ref{eq:HLF-d2}.}
\label{tab:rankHLF}
\end{table}

Structuring a matrix as shown in Figure \ref{hmatrix-fig} can lead to significant increase in maximum rank of the compressed matrix. Table-\ref{tab:rankHLF} shows the maximum possible rank of a $256\times 256$ matrix compressed to the same number of parameters using the two compression techniques - LMF and HLF. As shown, HLF can effectively double the rank of the matrix for the same number of parameters. To compress a matrix by a given compression factor, HLF has 2 different parameters, $j$ and $k$, to regulate the rank of the matrix. Hence, we see a range of rank values. Maximum rank is achieved when k=1. The value for j when k=1 can be calculated for different compression factors using equation \ref{eq:HLF-d2}.

Apart from the storage reduction, HLF also leads to a reduction in the number of computations. Assuming a batch size of 1 during inference, HLF leads to inference speed-up by using the associative property of matrix products to calculate the matrix-vector product - Algorithm~\ref{alg:hybrid-mvalg} shows how to calculate the matrix vector product when the matrix is represented using HLF. This algorithm avoids expanding the matrix $A'$, $B$ and $C$ into $A$.

Algorithm~\ref{alg:hybrid-mvalg} uses the associative property of matrix products to gain the computation speedup. For a matrix vector product between a matrix of size $m\times n$ and a vector of size $n\times1$, the number of operations required to compute the product is $m\times n$ ~\cite{linalgbook}. Referring to Algorithm~\ref{alg:hybrid-mvalg}, number of operations required to calculate O\textsubscript{1:j} is j$\times$n. The Temp1 variables need $k*n$ operations and calculating O\textsubscript{j+1:m} needs k*(m-j) operations. Thus, the reduction in number of operations when we use Algorithm~\ref{alg:hybrid-mvalg} is:
\begin{equation}
\label{eq:hybrid-acr}
\frac{ m \times n  }{j \times n + k \times n + k\times (m-j)}  
\end{equation}

\subsubsection{Impact on output feature vector}
\label{sec:outputfeature}
Algorithm \ref{alg:hybrid-mvalg} shows that, HLF divides the output into two stacked sub-vectors. One is a result of a fully-parameterized multiplication, $A'\times I$ (Line 1,  Algorithm \ref{alg:hybrid-mvalg}). The other is the result of the low rank multiplication : $B\times C \times I$ (Line 2-3,  Algorithm \ref{alg:hybrid-mvalg}). Thus, the upper sub-vector has ``richer'' features created from a higher dimensional embedding, while the lower sub vector has ``constrained'' features created from a lower dimensional embedding. By incorporating the HLF structure during training, we force an RNN to learn ``richer'' features in the upper sub-vector and the ``constrained'' features in the lower sub-vector. Because a RNN is followed by another RNN or a softmax layer, this restructuring should not impact the subsequent layers. \textit{Thus, HLF structure combines the assumptions A2 and A3 that were discussed previously}. 


\subsection{Why HLF leads to larger rank than LMF for same number of parameters?}
 HLF is an extension of LMF. To understand this, let us revisit Figure \ref{hmatrix-fig}, where the matrix 
 \begin{gather*}
    \boldsymbol{A} = [\boldsymbol{A'}\;;\; \boldsymbol{BC}] \\     
\end{gather*}
where $\boldsymbol{A} \in R^{m \times n}$, $\boldsymbol{A'} \in R^{j \times n}$, $\boldsymbol{B} \in R^{(m-j)\times k}$ and $\boldsymbol{C} \in R^{k \times n}$. Then we can rewrite the matrix as,
\begin{gather*}
    \boldsymbol{A} = [\boldsymbol{I}\;,\; \boldsymbol{0_1};\boldsymbol{0_2}\;,\;\boldsymbol{B}] [\boldsymbol{A'}\;;\;\boldsymbol{C}] 
\end{gather*}
where $\boldsymbol{I} \in R^{j \times j}$, $\boldsymbol{0_1} \in R^{j \times k}$ and $\boldsymbol{0_2} \in R^{(m-j)\times j}$. The above equation could be re-written as -
\begin{gather*}
    \boldsymbol{A} = \boldsymbol{U'}\boldsymbol{V'}
\end{gather*}
where $\boldsymbol{U'} \in R^{m \times (j+k)}$ and $\boldsymbol{V'} \in R^{(j+k) \times n}$. Both U' and V' can have a maximum rank  of $j+k$. The maximum value of this rank is achieved when k=1 and j is calculated as discussed in Table \ref{tab:rankHLF}. Let this value be $d$. A standard LMF decomposition of A will also lead to a representation of the form $\boldsymbol{UV}$, but this representation will have same parameters as HLF only if the rank of both $\boldsymbol{U}$ and $\boldsymbol{V}$ is at most $(d+1)/2$. Thus, HLF can be regarded as a $(d+1)$ ranked LMF of A, with a sparsity forcing mask that reduces the number of parameters to express the $(d+1)$ ranked matrix significantly. \textbf{This is why HLF can double the rank of the matrix when compared to an iso-parameter LMF matrix}. 
 
 Neural networks seldom learn structured sparsity unless they are forced to \cite{blockprune}, thus, an RNN trained with the LMF structure will rarely end up learning the same structure as HLF. \textbf{The pre-determined HLF structure effectively creates a sparsity forcing mask}. Such a sparsity forcing mask also leads to creation of the decoupled output feature vectors as described in section \ref{sec:outputfeature}.
 
 
\section{Results}
\label{sec:results}

\begin{figure*}
    \centering
    \begin{subfigure}{\textwidth}
        \centering
        \includegraphics[width=0.475\linewidth]{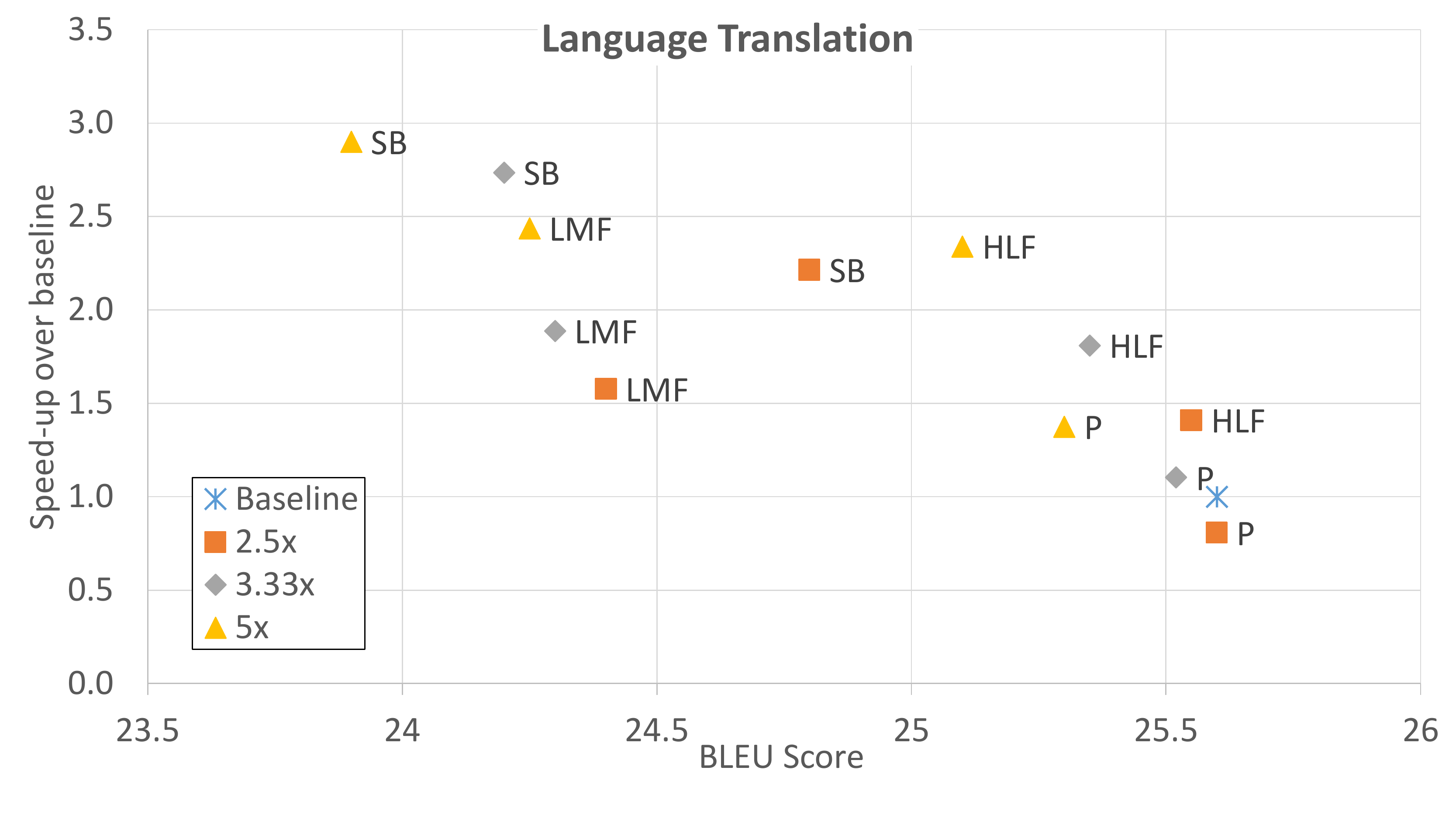}%
        \hfill
        \includegraphics[width=0.475\linewidth]{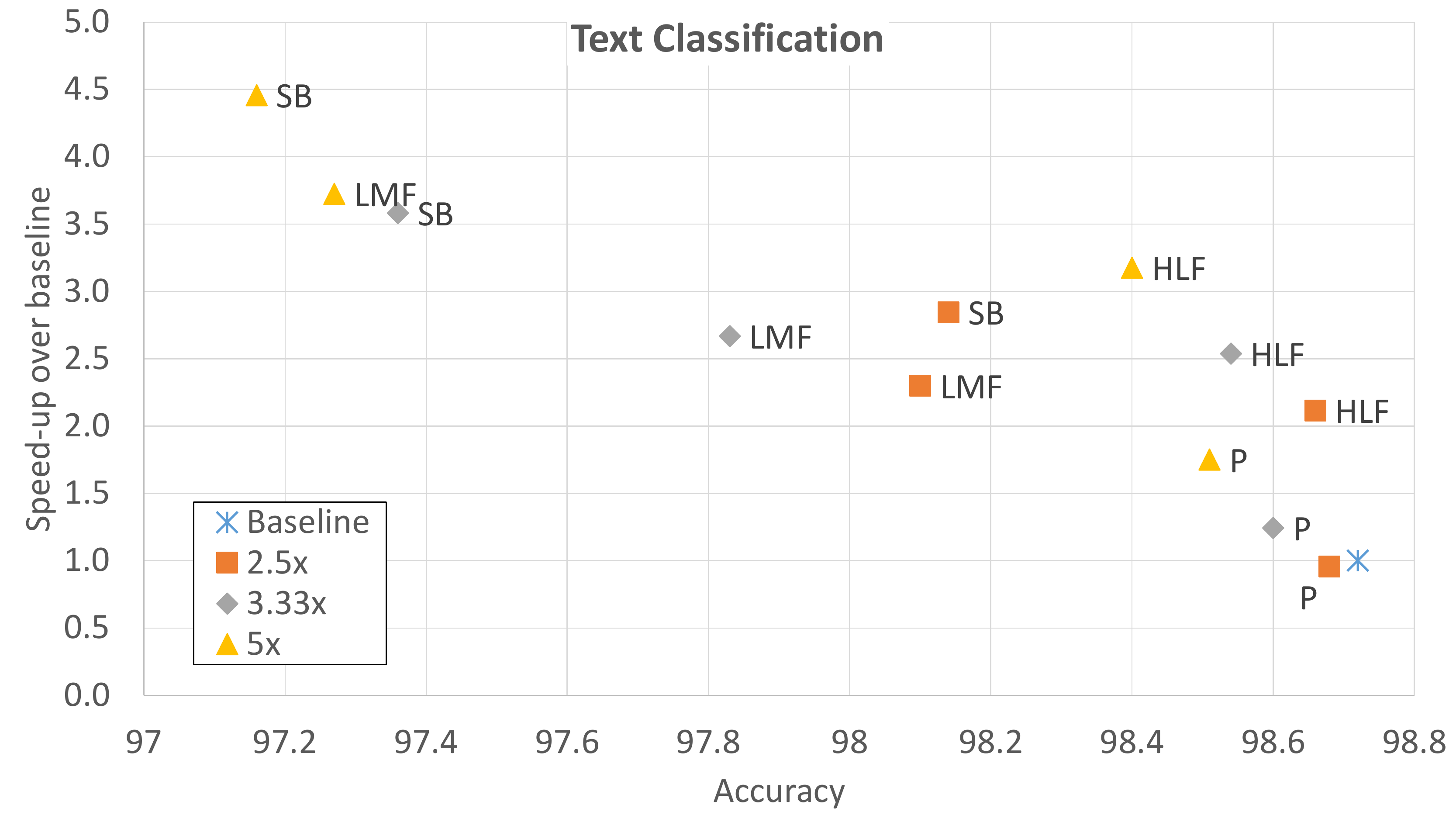}
        \caption{For both figures, HLF exists in the top-right corner, providing the best trade-off in terms of accuracy, compression and speed-up. For all 3 compression points, HLF provides better accuracy than LMF and SB and better run-time than pruning. \textit{\underline{(Left) Language Translation:}} \textbf{HLF improves the BLEU score achieved by LMF by 2.3\% to 4.5\% and by Small Baseline by 2.8\% to 4.1\%. At the same time, HLF improves the inference run-time over pruning by $1.5-1.74\times$}. \textit{\underline{(Right) Text Classification:}} \textbf{HLF can improve the accuracy achieved by LMF by up-to 1.2\% and SB by up-to 1.3\% and improve the run-time achieved by pruning by up-to $1.2\times$}}
        \label{fig:nmt}
        \label{fig:textclass}
    \end{subfigure}
    \vskip\baselineskip
    \begin{subfigure}{\textwidth}
        \centering
        \includegraphics[width=0.475\linewidth]{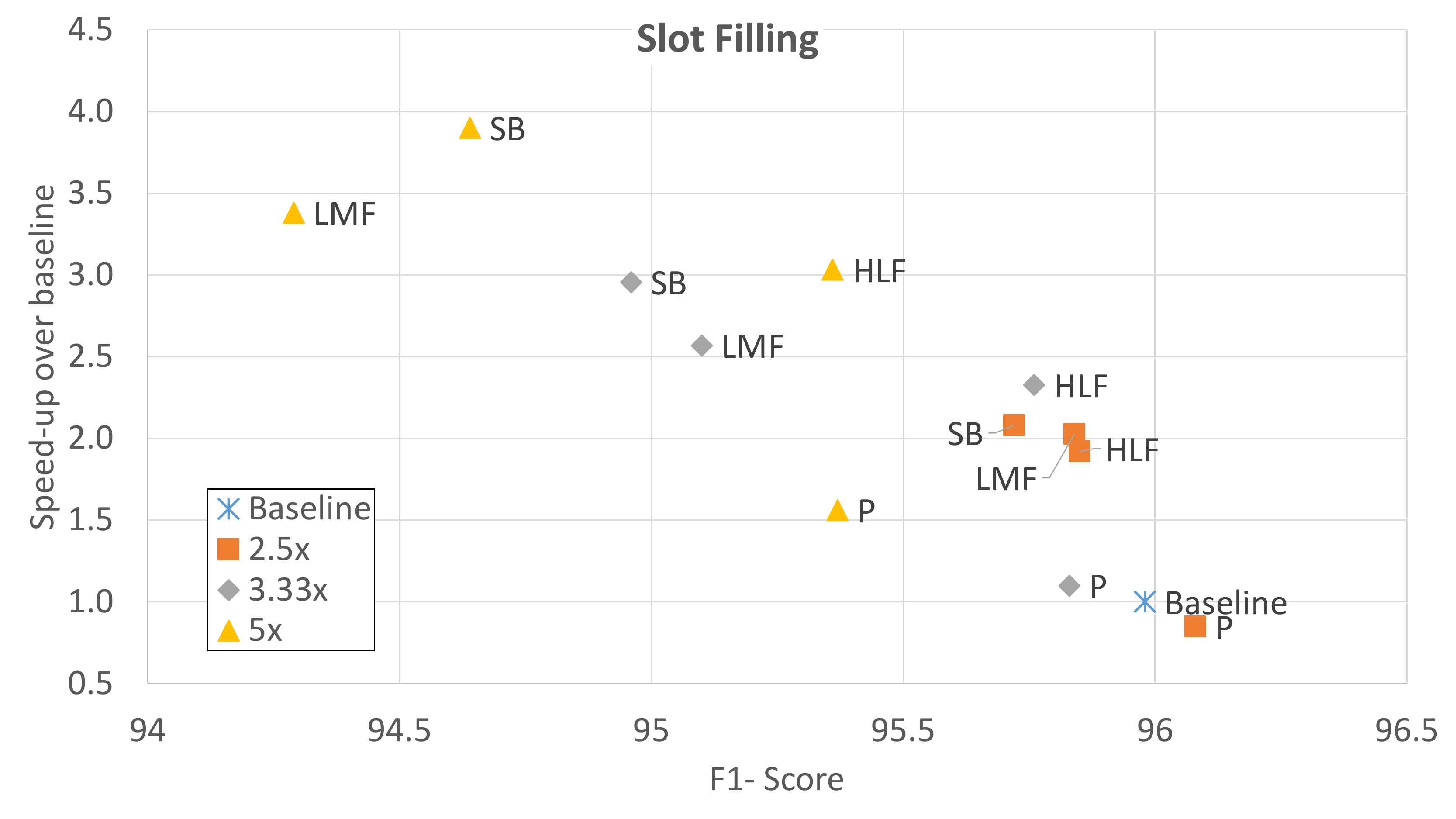}%
        \hfill
        \includegraphics[width=0.475\linewidth]{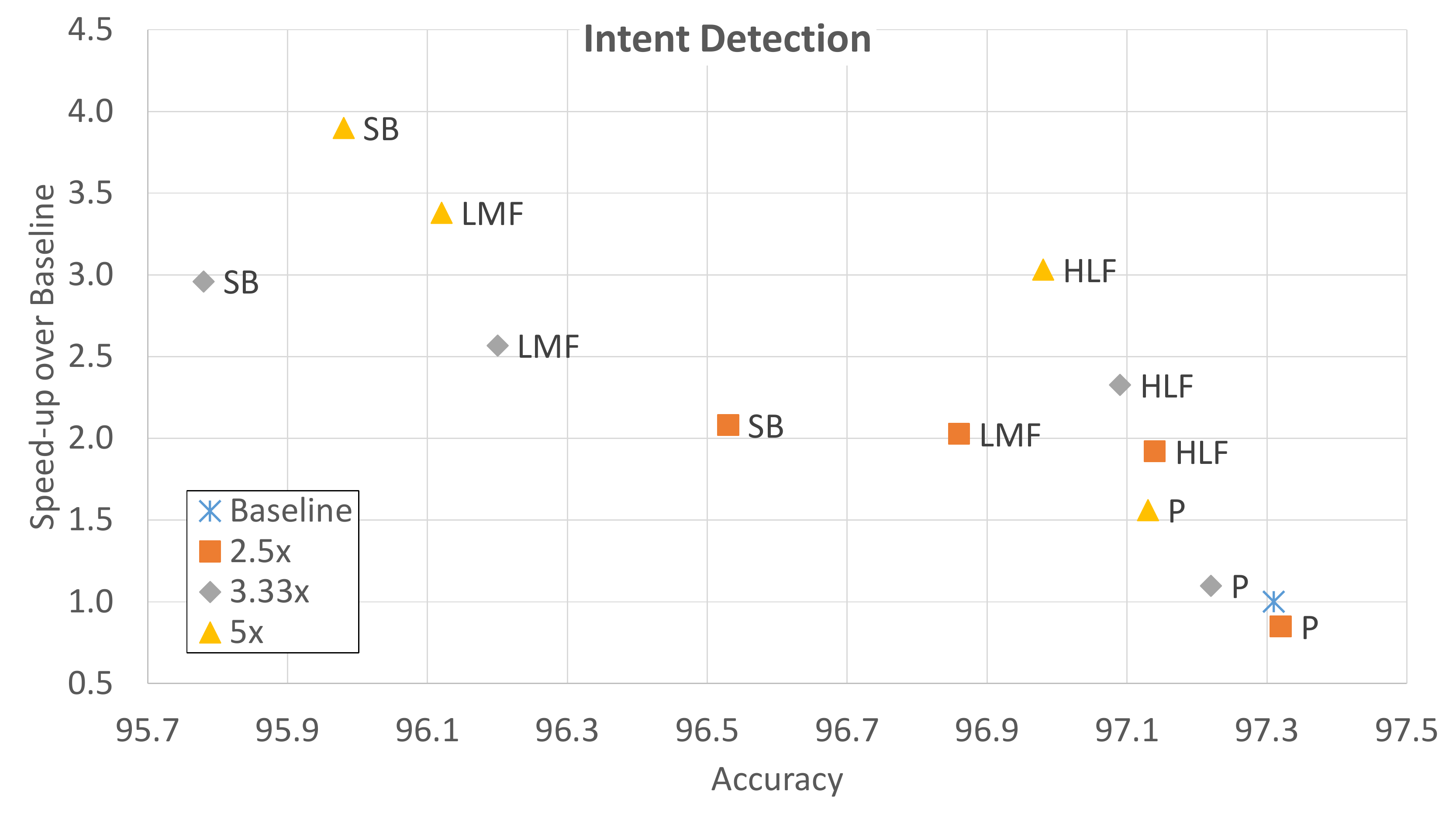}
        \caption{For both figures, HLF exists in the top-right corner, providing the best trade-off in terms of accuracy, compression and speed-up. \textit{\underline{(Left) Slot Filling}}: For all 3 compression points, HLF provides better accuracy than LMF and SB and better run-time than pruning. \textbf{HLF can improve the accuracy achieved by LMF and SB by up-to 1.2\% and improve the run-time achieved by pruning by up-to $1.26\times$}. \textit{\underline{(Right) Intent Detection}}: For all 3 compression points, HLF provides better accuracy than LMF and SB and better run-time than pruning. \textbf{HLF can improve the accuracy achieved by LMF and SB by up-to 1\% and improve the run-time achieved by pruning by up-to $1.26\times$}}
        \label{fig:intent}
        \label{fig:slot}
    \end{subfigure}
    \begin{subfigure}{\textwidth}
    \centering
    \includegraphics[width=0.475\textwidth]{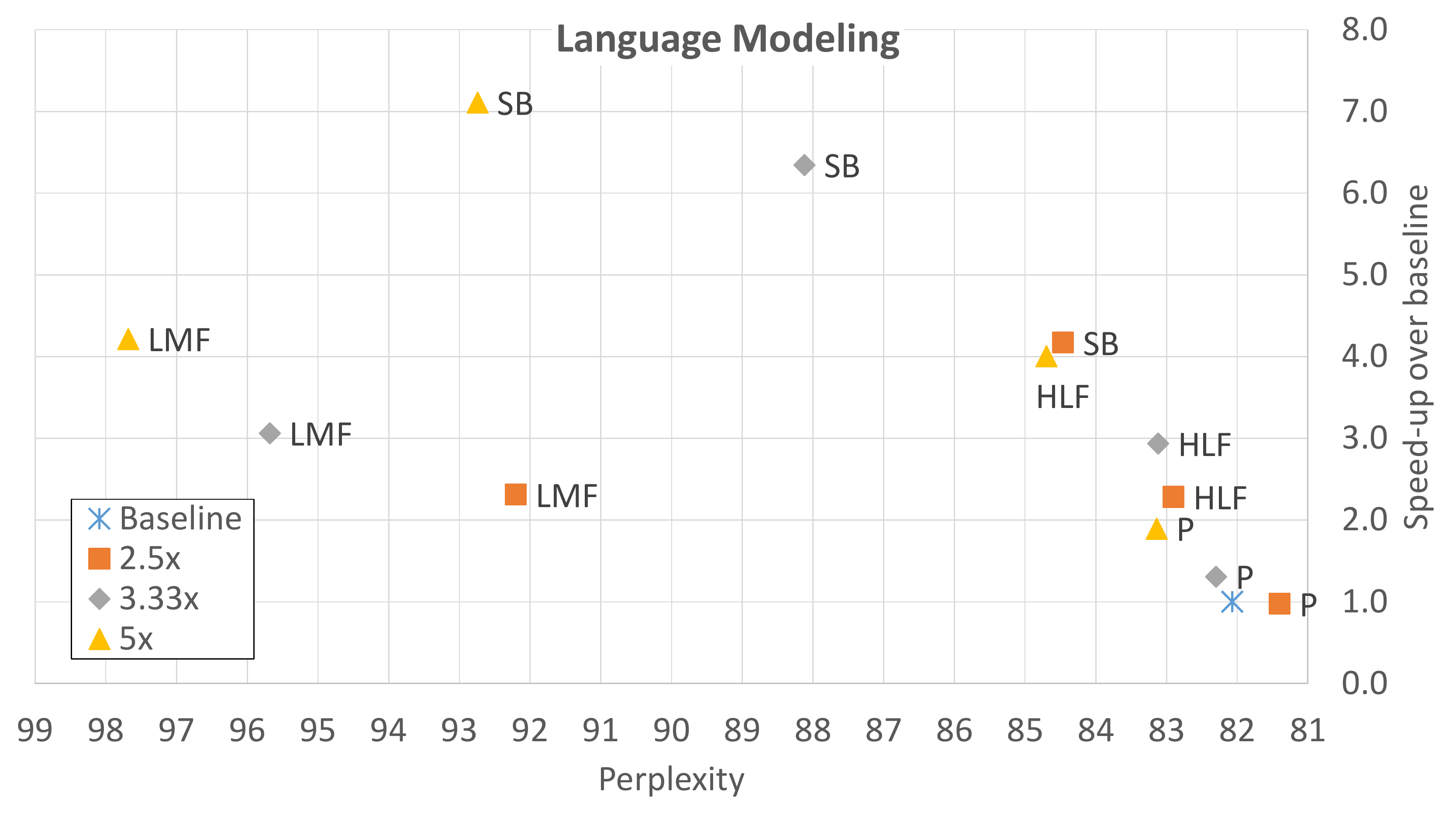}
    \caption{Language Modeling: For LM, lower values of perplexity are better. HLF provides a viable alternative to both pruning and LMF at compression factors of 2.5x, 3.0x and 5x. \textbf{HLF can achieve 17\% better perplexity than LMF, 9\% better perplexity than small baseline and $2.32\times$ better inference run-time than pruning.}}
\label{fig:ptblm_med}
\end{subfigure}
\caption{Speed-up over baseline vs Accuracy comparing the baseline with a smaller baseline and the baseline compressed using different compression schemes at varying compression factors. Speed-up values $>1$ indicate a decrease in inference run-time and values $<1$ indicate an increase in inference run-time. \textbf{For each compression factor, the compression scheme that is most to the top-right is the ideal choice}. \textit{In case of perplexity, lower values are better}. Thus, the graphs are plotted in a slightly different way to still adhere to the fact that the most ideal choice of compression is in the top-right corner. P = Pruning, LMF = Low rank matrix factorization, HLF = Hybrid matrix decomposition, SB = Smaller baseline. \textbf{The best way to view this figure is to either focus on a compression point and see how the Pareto curve of speed-up vs accuracy changes as we add HLF or focus on an accuracy region and see what compression schemes provides the best run-time at highest compression factor.}}
\end{figure*}

We compare HLF with LMF and 3 other compression techniques -- model pruning, small baseline and a structured matrix based technique called block circular decomposition. 
These techniques and why they need to be considered are discussed below: 
\begin{itemize}
    \item Pruning: Model pruning~\cite{suyog} induces sparsity in the matrices of a neural network, thereby reducing the number of non-zero valued parameters that need to be stored. Pruning creates sparse matrices which are stored in a specialized sparse data structure such as CSR.  The overhead of traversing these data structures while performing the matrix-vector multiplication can lead to poorer inference run-time than when executing the baseline, non-sparse network. \textit{Thus, while pruning is an effective compression technique, its run-time performance on CPUs can make it a less appealing choice for compression}. We use the magnitude pruning framework provided by \cite{suyog}. While there are other possible ways to prune, recent work \cite{bestprune} has suggested that magnitude pruning provides state-of-the-art or comparable performance when compared to other pruning techniques \cite{varprune,l0prune}. 
    \item Small Baseline: Additionally, we train a smaller baseline with the number of parameters equal to that of the compressed baseline. This serves as a useful point of comparison because of two reasons. 
    \begin{itemize}
        \item  First, to check if compression of a larger network leads to better accuracy than compressing a network by reducing its dimensions (size of hidden layer or number of layers). \textbf{This can help us verify if the network was originally over-parameterized}. 
        \item Second, \textbf{to establish the hypothesis whether HLF's creation of a stacked output feature vector as described in section \ref{sec:outputfeature} adds any useful information in the network}. Smaller baseline creates output feature vector that is created from a high-dimensional embedding only. 
        HLF, additionally concatenates the output features created from lower dimensional embedding. Thus, comparing the accuracy of HLF with Smaller baseline helps evaluate the usefulness of the output features created using lower dimensional embedding. 
    \end{itemize}
    \textbf{Given the significant slow-down of inference of BCD compressed networks, we do not discuss the results sing BCD compression in the rest of the paper}. 
\end{itemize}

\subsection{Experiment Setup}
\textbf{Measuring inference run-time:} In order to compare the inference run-time of RNN cells compressed using pruning, LMF and HLF, we implemented these cells in C++ using the Eigen library. This paper focuses on inference on an edge device. As a result, we make the assumption that the batch size of the application will be 1 while measuring the run-time of an application. However, the observations regarding run-time should remain consistent for larger batch sizes as well.  We ran our experiments on a single cortex-A73 core  of the Hikey 960 board. The size of L3 cache is 2MB.

\textbf{Training infrastructure:} We use Tensorflow 1.14 to train our networks on a cluster of 2 RTX 2080 Ti NVidia GPUs with 11 GB Memory. The training settings for the benchmarks evaluated can be found in the reference paper for each benchmark. 

\textbf{What do we compress?} We compress the LSTM/GRU layers in each application. We do not compress the embedding layers using HLF. 

The amount of compression determines the rank of the compressed matrices when we use LMF to compress an NLP application and the sparsity of the pruned matrix when we use pruning as our choice of compression technique. Similar to LMF, the amount of compression determines the rank of the compressed matrices when we use HLF as our choice of compression. However, for HLF two parameters j and k control the rank of the matrix. We use a sweep starting with k=1 to determine the exact values of j and k that help us achieve a good accuracy.

\textbf{What do we compare?}: We compare the accuracy and inference run-time of all compression technologies at iso-compression factors for various compression techniques.

\subsection{Comparison of compression techniques across different ML tasks}
\label{sec:comparison}

The impact of compression on accuracy is compared for 5 benchmarks -- machine translation, natural language understanding (intent detection and slot filling), text classification and Language Modeling. These tasks are some of the most important NLP applications that run on edge and embedded devices like smart phones, smart watches and smart homes. 

\subsubsection{Language Translation}
\label{sec:nmt}
We use the English to Vietnamese translation model in \cite{nmt_vien}. The model uses 2-layer LSTMs of size 512 units with bidirectional encoder (i.e., 1 bidirectional layers for the encoder), embedding of dim 512 and an attention layer. We used the hyper-parameters in \cite{nmt_vien} to train the network while modifying the learning rate values used. We sweep the learning rates values from $\times 0.1$ to $3.0$ in multiples of 3. For HLF, we used the value of k=2.

Figure ~\ref{fig:nmt} shows the results of compressing the LSTM layers in the NMT\_VIEN baseline by $2.5\times$, $3.33\times$ and $5\times$. \textbf{HLF improves the BLEU score achieved by LMF by 2.3\% to 4.5\% and by Small Baseline by 2.8\% to 4.1\%. At the same time, HLF improves the inference run-time over pruning by $1.5\times - 1.74\times$}. 

\subsubsection{Language Modeling}
\label{sec:ptblm}
We use the medium LM model from ~\cite{ptblm} as our baseline. The PTB (Medium) baseline has 2 LSTM layers each with a hidden vector of size 650 with a vocabulary size of 10,000 words from the English vocabulary. We used the hyper-parameters in \cite{ptblm} and train the compressed networks for 50 more epochs than in baseline. The baseline network is trained for 39 epochs. For the first 6 epochs the learning rate used is of value 1, and after that we decrease it by a factor of 1.2 after each epoch. We clip the norm of the gradients at 5 and use dropout of value 0.35. For HLF, we used the value of k=4.

Figure ~\ref{fig:ptblm_med} shows the results of compressing the LSTM layers in the PTB (Medium) baseline by $2.5\times$, $3.33\times$ and $5\times$.  Lower the perplexity, better the model. Pruning achieves the same (sometimes better) perplexity than baseline and other compression techniques.  LMF leads to significant loss in perplexity for all compression factors while HLF achieves better perplexity than LMF and faster inference run-time than baseline and pruned networks for all compression factors. \textbf{In fact, HLF can achieve 17\% better perplexity than LMF, 9\% better perplexity than small baseline and $2.32\times$ better inference run-time than pruning}. The preferred choice of compression scheme for different compression factors will depend on whether slight loss in perplexity can be accommodated for faster inference run-time or not. However, HLF still manages to serve as a more viable alternative to pruning than LMF for inference on edge CPUs.

\subsubsection{Text Classification}
\label{sec:textclass}
We use the text classification network in \cite{textclass} evaluated on the SemEval-2010 dataset. The baseline network has 1 bidirectional LSTM layers with hidden vector of size 256.  We used the hyper-parameters in \cite{textclass} to train the baseline and as the initial hyperparameters explored for the compressed networks. We trained the compressed networks for additional 20 epochs while exploring learning rates of $10\times$ and $1/10$ than the baseline learning rate.  The baseline model was trained using AdaDelta with a learning rate of 1.0. The model parameters  were regularized with L2 regularization strength of $10^-5$.

Figure \ref{fig:slot} shows the results for compressing the text classification network by $2.5\times - 5\times$. \textbf{HLF can improve the accuracy achieved by LMF by up-to 1.2\%, by SB by up-to 1.3\% and improve the run-time achieved by pruning by almost $1.20\times$.} 

\subsubsection{Intent Detection and Slot Filling}
\label{sec:intSlot}
We used the benchmark published in \cite{slotintent}. This benchmark is trained on the ATIS dataset and jointly trains for intent detection and slot filling. The benchmark uses 1 LSTM layer of size 128 along with attention layers. For HLF, we used the value of k=1.    

Figure \ref{fig:slot} shows the results for slot filling task. \textbf{HLF can improve the F1-accuracy achieved by LMF and SB by up-to 1.2\% and improve the run-time achieved by pruning by up-to $1.26\times$}. Figure \ref{fig:intent} shows the result for the intent classification task. Due to joint training, the network used for slot filling and intent classification is the same. As a result, the runtime improvement of HLF over pruning is exactly the same as for the slot filling task. \textbf{Additionally, HLF improves the intent classification accuracy by up to 1\% over LMF and small baseline.} 

\subsection{Ablation Studies}
\subsubsection{Compressing Word Embedding layers using HLF}
We compressed the input word-embedding layers in the PTB-LM model discussed in section \ref{sec:ptblm}, without compressing other layers in the network. However, even 3$\times$ compression using HLF led to 8\% loss in perplexity score. 

\subsubsection{Orthogonality of word embedding compression methods and HLF} 
We ran experiments where we prune the word embedding layers in the PTB-LM model in section \ref{sec:ptblm} by $2\times$ while  keeping the LSTM layers uncompressed, leading to 83.1 perplexity score. We were able to further compress this network by $2\times$ using HLF with only 1 point loss in perplexity score, indicating that HLF is compatible with techniques used for compressing word-embedding layers.

\section{Discussion}
\label{sec:discussion}

Effectively, HLF acts as an alternative to LMF whenever compression using pruning does not lead to the required run-time benefit and LMF leads to loss in accuracy. HLF has a better accuracy than LMF for most evaluation points, validating the assumption in the paper that rank of a matrix in a RNN is important for better task accuracy in NLP applications. Additionally, HLF has a better accuracy than smaller baseline. This validates the assumption of the importance of constrained features in addition to the richer features in a Small Baseline network.
\section{Limitations}
While HMD provides significant benefits over LMF, there are two limitations associated with the technique:
\begin{itemize}
    \item  The unique nature of RNNs (Assumption A1-A3) makes HLF a natural fit for LSTM/GRU layers. However, these assumptions are not valid for the final classification layer. In classification layer, HLF will lead to more expressive output for certain classes (the top part of HLF matrix) in the dataset and less expressive output for the rest of the classes (bottom part of HLF matrix). 
    \item HMD is a training aware compression and cannot be applied to a pre-trained network. 
\end{itemize}
\section{Conclusion}
Choosing the right compression technique requires looking at three criteria -- compression factor, accuracy, and run-time. Pruning is an effective compression technique, but can sacrifice speedup over baseline for certain compression factors. LMF achieves better speedup than baseline for all compression factors, but leads to accuracy degradation. This paper introduces a new compression scheme called HLF, which preserves the dense structures of LMF while effectively doubling the rank of the matrix using an intelligent structure by design. This leads to $2\times$ faster inference run-time than pruning and up-to $16\%$ better accuracy than LMF.

\bibliographystyle{acl_natbib}
\bibliography{main}
\end{document}